\documentclass[runningheads]{llncs}
\usepackage[T1]{fontenc}
\usepackage{graphicx}
\usepackage{booktabs}
\usepackage{amsmath}
\usepackage{amsfonts}
\usepackage{multirow}
\usepackage{hyperref}
\usepackage{xcolor}
\usepackage{lipsum}
\usepackage{marvosym}

\newcommand{\thepoint}{R\?.\?}
\newcounter{point}

\begin{document}
\title{Gaze-DETR: Using Expert Gaze to Reduce False Positives in Vulvovaginal Candidiasis Screening}

\author{
Yan Kong\inst{1}$^{\dag}$ \and
Sheng Wang\inst{1,2}$^{\dag}$ \and
Jiangdong Cai\inst{1} \and
Zihao Zhao\inst{1} \and \\
Zhenrong Shen\inst{2} \and
Yonghao Li\inst{1} \and
Manman Fei\inst{2} \and \\
Qian Wang\inst{1,3}$^{(\textrm{\Letter})}$
}
\authorrunning{Y. Kong, S. Wang et al.}

\institute{
School of Biomedical Engineering \& State Key Laboratory of Advanced Medical Materials and Devices, ShanghaiTech University, Shanghai, China \and
School of Biomedical Engineering, Shanghai Jiao Tong University, Shanghai, China \and
 Shanghai Clinical Research and Trial Center, Shanghai, China
\email{qianwang@shanghaitech.edu.cn} }

\renewcommand{\thefootnote}{}
\footnotetext{${\dag}$ Yan Kong and Sheng Wang contributed equally to this study.}

\maketitle
\begin{abstract}
Accurate detection of vulvovaginal candidiasis is critical for women's health, yet its sparse distribution and visually ambiguous characteristics pose significant challenges for accurate identification by pathologists and neural networks alike. 
Our eye-tracking data reveals that areas garnering sustained attention - yet not marked by experts after deliberation - are often aligned with false positives of neural networks. 
Leveraging this finding, we introduce Gaze-DETR, a pioneering method that integrates gaze data to enhance neural network precision by diminishing false positives. 
Gaze-DETR incorporates a universal gaze-guided warm-up protocol applicable across various detection methods and a gaze-guided rectification strategy specifically designed for DETR-based models. 
Our comprehensive tests confirm that Gaze-DETR surpasses existing leading methods, showcasing remarkable improvements in detection accuracy and generalizability.

\keywords{Candida Detection, DETR, Eye-tracking}
\end{abstract}

\section{Introduction}
Vulvovaginal candidiasis is a prevalent fungal infection caused by candida.
It has been estimated that about 75\% of women will experience vulvovaginal candidiasis at least once in their lifetime, leading to significant discomfort and healthcare resource usage~\cite{gonccalves2016vulvovaginal,benedict2019estimation,willems2020vulvovaginal}. 
Thin-prep cytologic test (TCT) plays a vital role to detect vulvovaginal candidiasis. 
However, visually analyzing TCT images requires intensive labor from pathologists, thus limiting the efficiency and scalability of cervical screening~\cite{koss1989papanicolaou,cai2023progressive}. 
Therefore, there is an urgent need for a computer-assisted diagnosis (CAD) system capable of analyzing TCT images automatically. 

\begin{figure}[t]
    \centering
    \includegraphics[width=\textwidth]{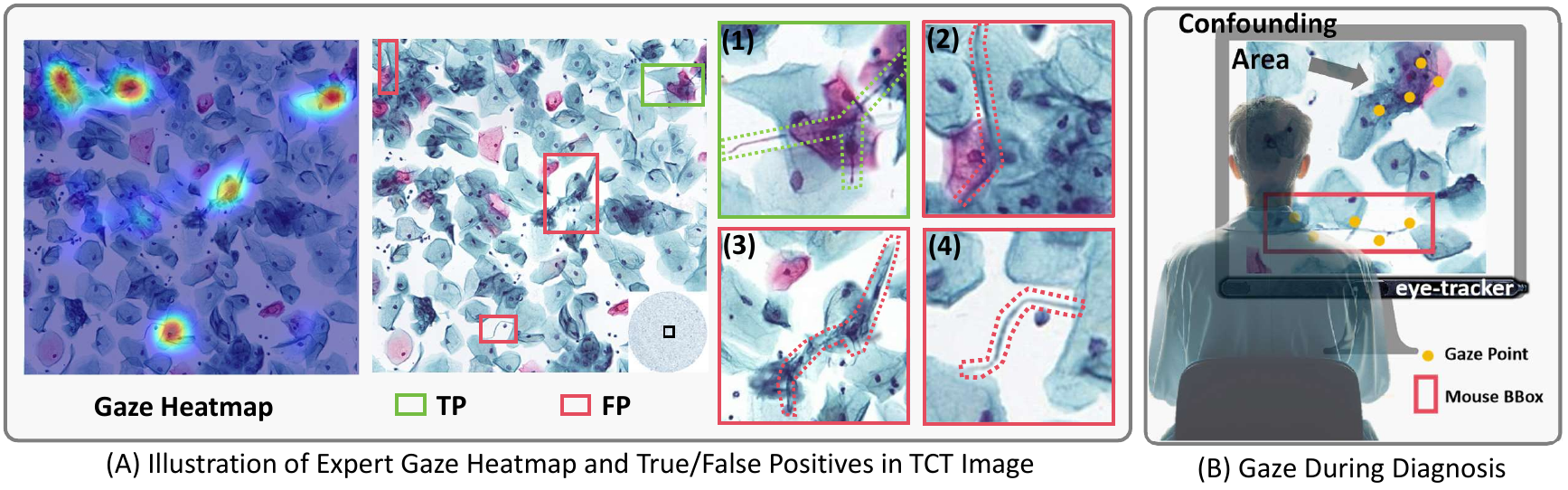}
    \caption{
        (A) TCT image samples displaying false positives (red) versus true positives (green), and corresponding gaze heatmap. (B) Gaze tracking during labeling: An expert's gaze (yellow) and mouse-labeled regions on a TCT image. Regions where experts  carefully review without annotating are usually confounding area.
    }
    \label{fig_intro}
\end{figure}

In recent years, the emergence of deep-learning-based CAD systems has revolutionized cervical screening by facilitating automated detection of cervical abnormalities~\cite{rahaman2020survey}. 
Despite notable progress, the existing methods still struggle with the precise identification of candida.
The reasons include \textbf{a) the imbalance between candida and background cells}, and \textbf{b) the thin, faint appearance and frequent occlusions} as shown in Fig. \ref{fig_intro} (A).
These factors hinder the model from distinguishing candida from similar cellular structures and contaminants, leading to a notable increase in false-positive results.
An example TCT image is illustrated in Fig. \ref{fig_intro} (A), in which one true positive (green) and three false positives (red) with sharp cell structures and linear contaminants are displayed. It can be observed that RetinaNet~\cite{lin2017focal} mistakenly identifies the three of four suspected regions as candida.

To address the issue of high false positive rates in TCT image analysis, we leverage the eye movement patterns of pathologists collected by eye-trackers during their image reading. 
The gaze data, which is long valued for enhancing diagnostic accuracy and understanding human error~\cite{wu2019eye,brunye2019review,voisin2013predicting}, has recently been shown to strengthen various CAD tasks~\cite{wang2022follow,ma2023eye,ji2023mammo,wang2023crafting,zhao2024mining,wang2023learning,wang2024gazegnn,wang2023gazesam}. 
By studying the connection between gaze patterns and detection errors, we observe a correlation between prolonged gaze on candida-like objects and the likelihood of those areas being identified as false positives by detection models.
This insight, illustrated by a comparison between the detections of RetinaNet~\cite{lin2017focal} and the expert gaze heatmap in Fig. \ref{fig_intro} (A), enables the strategic use of expert annotations and eye-tracking data to identify and correct potential false positives during model training.
Note that the gaze data is passively collected when pathologists conduct manual labeling of candida boxes, causing no disruption to pathologists as depicted in Fig. \ref{fig_intro} (B).

With all the investigation, we propose \textbf{Gaze}-guided \textbf{De}tection \textbf{Tr}ansformer (Gaze-DETR), an innovative training strategy that incorporates expert gaze to enhance the ability of detectors to distinguish candida from false positives. 
Our model is built upon DETR~\cite{carion2020end} with a two-step training strategy comprising \textit{gaze-guided warm-up} followed by \textit{gaze-guided rectification}.
Gaze-guided warm-up is employed to additionally predict the `gaze only' category, highlighting regions where experts devote visual attention to carefully review yet without annotating them as actual candida instances. 
This warm-up strategy allows the detector to gain additional insights into confounding candida-like instances, thereby addressing the issue of imbalanced candida quantities. 
Subsequently, gaze-guided rectification is employed to construct more focused queries in the vicinity of confounding areas. 
These queries are then rectified by our model, which can be considered as a form of hard sample mining, effectively resolving the challenge of distinguishing difficult candida instances. 
The gaze information is not required during inference.
The extensive experiments demonstrate the effectiveness of our approach in improving candida detection performance and its compatibility with various kinds of detection models. 


\section{Method}

\begin{figure}[t]
    \centering
    \includegraphics[width=0.99\linewidth]{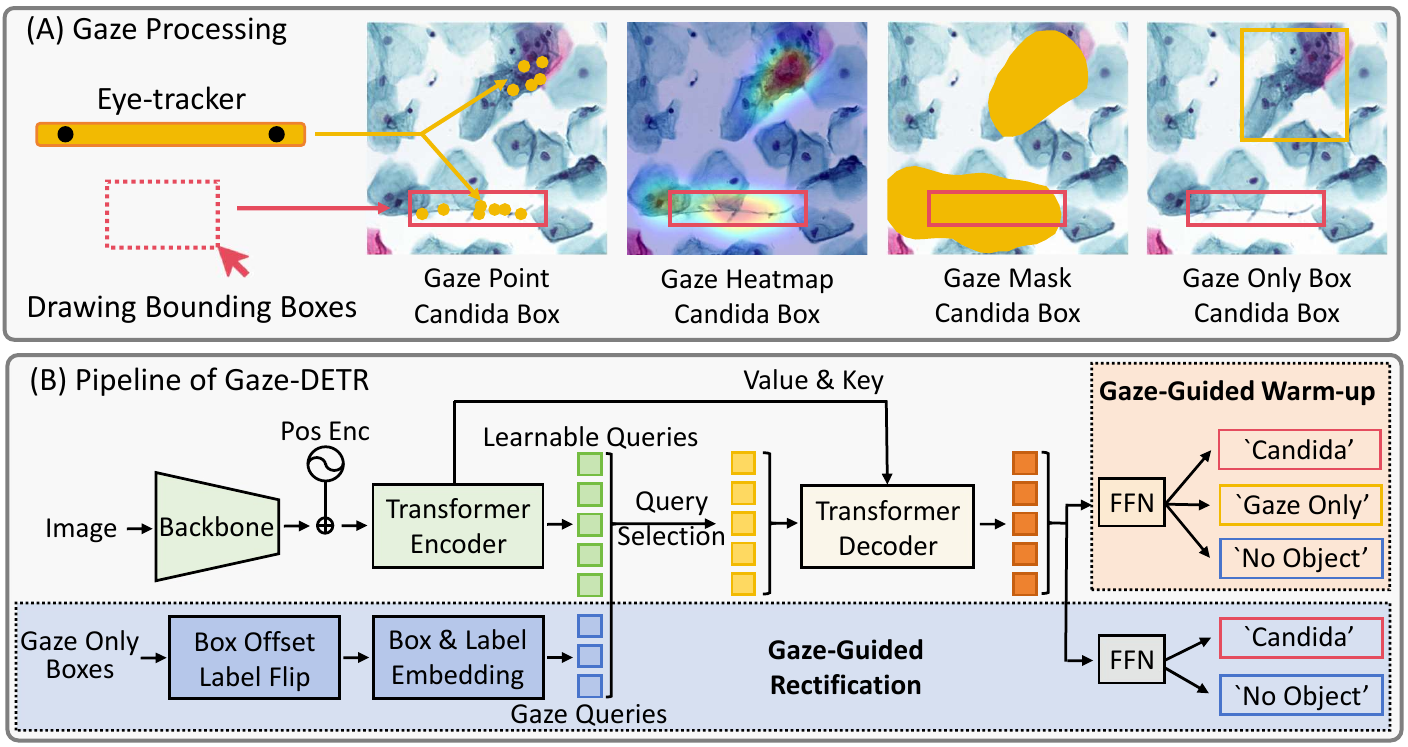}
    \caption{Overview of the Gaze-DETR. (A) Illustration of Gaze Processing, which derives gaze points to `gaze only' boxes. (B) Pipeline of the Gaze-DETR model, which integrates the gaze guidance into the DETR framework for candida detection.}
    \label{fig:gaze-predict}
\end{figure}

In this section, we start with the processing of gaze data, introducing the definition of `gaze only' boxes along with their underlying semantics. Then, we delve into the pipeline of our proposed Gaze-DETR, which incorporates expert knowledge within 'gaze only' boxes via gaze-guided warm-up and gaze-guided rectification. By leveraging gaze data, we aim to align the recognition capabilities of DETR-like detectors with those of human experts, enhancing the overall performance of our Gaze-DETR model.

\subsection{Gaze Processing}
To mine semantics within gaze data that are beneficial to mitigate false negatives, we introduce the definition of `gaze only' boxes and present its process pipeline in  Fig.~\ref{fig:gaze-predict} (A).
In this study, \textbf{`gaze only' boxes refer to regions receiving long-lasting attention from pathologists but are finally diagnosed as normal.}
These 'gaze only' boxes are typically associated with instances (e.g. cellular folds and thin contaminants in Fig.~\ref{fig_intro} (A)) that are challenging to discriminate in clinical practice, making them valuable for aligning DETR-like models with human recognition capabilities. 
To craft `gaze only' boxes, a Gaussian kernel is first applied to those spatiotemporally discretized gaze points, generating a heatmap~\cite{wang2022follow}. 
A threshold is then applied to this heatmap to identify regions of interest.
Subsequently, the small-sized regions are eliminated, and gaze boxes are extracted by detecting contours within the remaining regions. 
Finally, the final `gaze only' boxes are obtained by selecting the portions of the gaze boxes that do not contain candida boxes.

\subsection{Gaze-Guided Warm-up}

Compared with densely distributed cervical cells, the scarcity and sporadic distribution of candida in TCT images create an imbalance, which potentially leads to an unstable training phase. To overcome this, we propose a gaze-guided warm-up strategy for DETR-like detection models.

As shown in Fig.~\ref{fig:gaze-predict} (B), DETR extracts features through a backbone and further forwards them through the Transformer encoder to obtain refined features.
Unless otherwise specified, we follow the architecture of DAB-DETR~\cite{liu2022dab} instead of the original DETR~\cite{carion2020end} throughout this study. Hence, learnable queries are learned from refind features, instead of random initialized and self-updated. 
These learnable queries are then processed by the Transformer decoder, finally obtaining predictions via the shared feed-forward networks (FFNs). To stabilize its training,
 we expand the FFN's classification capability beyond the `candida' and `no object' categories to also include a `gaze only' category during the initial training phase. This strategy is plug-and-play across both DETR, and other CNN-based models as we will demonstrate in Section~\ref{sec:compat}.
While gaze only boxes are not actual candida, our strategy can 1) help the shallower layers of neural network learn to look at the discriminative appearance that looks like a candida.
2) prevent the deeper layers' gradient dilution when bias towards the majority (no object), which creates training instability.

\subsection{Gaze-Guided Rectification}
\begin{figure}[t]
    \centering
    \includegraphics[width=0.95\linewidth]{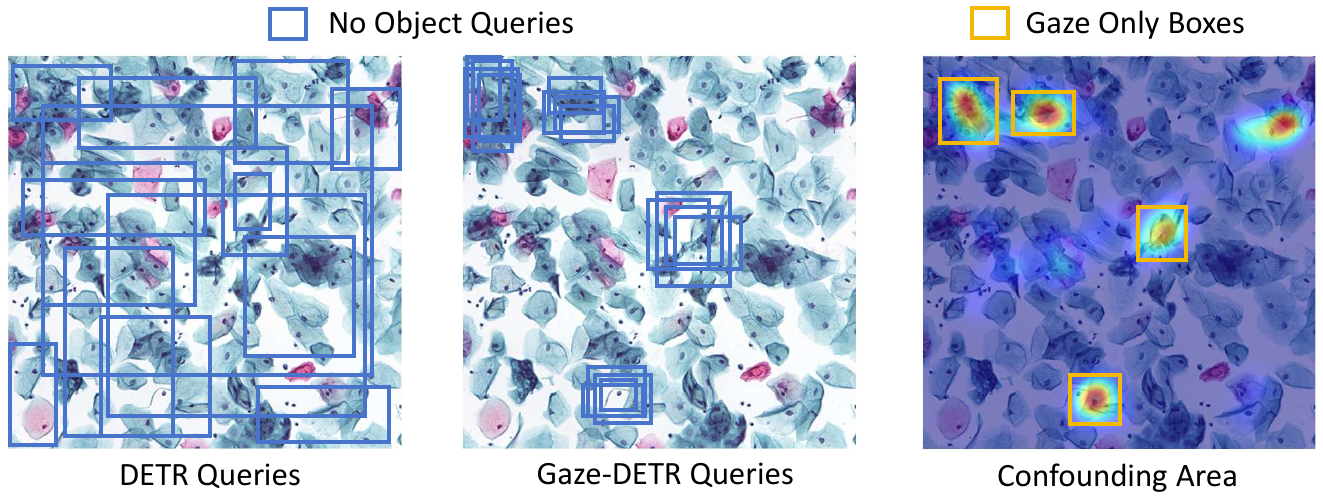}
    \caption{Illustration to compare ‘no object’ queries of standard DETR against Gaze-DETR. Gaze-DETR effectively leverages `gaze only' boxes to concentrate `no object' queries on confounding regions, compared to the nearly random distribution of `no object' queries of DETR.
     }
    \label{fig:query}
\end{figure}

The matching strategy for `no object' queries in DETR training significantly impacts model performance, especially in our task where a typical TCT image patch may contain only one or two instances of candida~\cite{zhang2022dino}. 
To enhance the model's focus on difficult areas, we introduce the strategy of gaze-guided rectification. 
This approach leverages gaze data to guide `no object' queries towards confounding areas, providing targeted supervision during training and improving the model's discriminative ability.
Fig. \ref{fig:query} contrasts the random distribution of `no object' queries in standard DETR with our Gaze-DETR, which concentrates queries on expert-identified regions.

To guide the detector's attention toward confounding areas, we introduce gaze queries, the position and content of which are both optimized by gaze data.
Due to eye-tracking device errors and the gaze processing method, the generated `gaze only' boxes tend to be offset from the corresponding objects, 
where the offset (\(\Delta x\), \(\Delta y\), \(\Delta w\), \(\Delta h\)) follows respective Gaussian distributions.
Thus, to make `gaze only' boxes better reflect the confounding areas, we replicate the `gaze only' boxes multiple times (four times in our model), and then utilize an explicit query position construction method, inspired by DAB-DETR~\cite{liu2022dab}, to offset the gaze boxes from (x$'$, y$'$, w$'$, h$'$) to (x$'$+\(\epsilon_x\), y$'$+\(\epsilon_y\), w$'$+\(\epsilon_w\), h$'$+\(\epsilon_x\)), where \(\epsilon_x\), \(\epsilon_y\), \(\epsilon_w\), \(\epsilon_h\) are random number following the corresponding Gaussian distribution.
Additionally, we explicitly flip the class of the gaze boxes to candida, drawing inspiration from DN-DETR~\cite{li2022dn}. 
The class flip ensures that the class labels of the queries are corrected during processing in the FFNs, further prompting the model to match false positives to the `no object' category.
To maintain consistent token dimensions for the input of the decoder, we introduce a query selection mechanism. 
The mechanism prioritizes all gaze queries and selects the learnable queries with the most important features learned from the decoder as the rest.
This mechanism enables the queries to concentrate on the confounding areas and candidate candida boxes to be retained.

\section{Experimental Results}

\begin{figure}[t]
    \centering
    \includegraphics[width=0.95\textwidth]{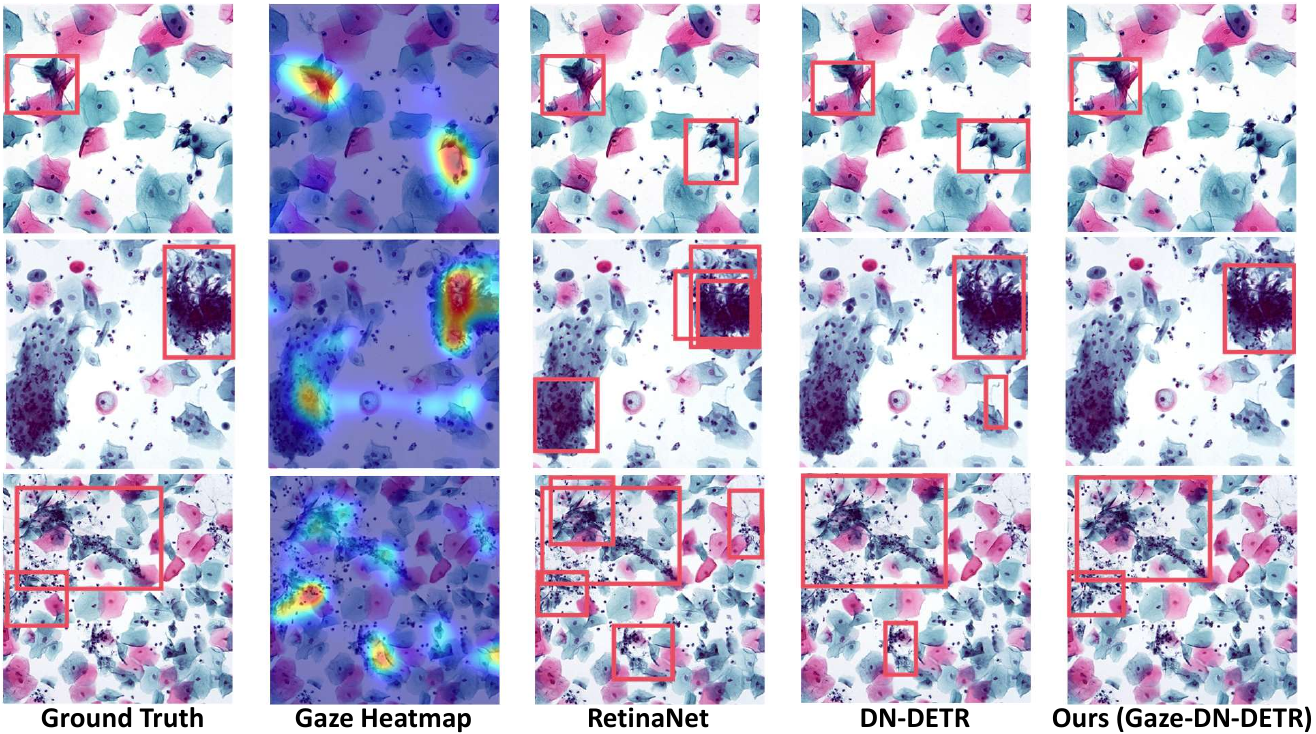}
    \caption{
        Comparison of qualitative detection results. 
        Our method significantly reduces false positive errors in areas that are carefully reviewed by experts, which are often prone to misclassifications by both RetinaNet and DN-DETR.
    }
    \label{fig_result}
\end{figure}
\subsection{Data and Experimental Settings}
\textbf{Image Collection} Our in-house dataset consists of whole slide images from patients with candida, which are cropped into patches of $1024\times1024$. 
These patches are meticulously examined and labeled by an experienced pathologist, who annotates candida with bounding boxes, resulting in a collection of 450 patches with 612 bounding boxes.
The dataset is further divided into training, validation, and test sets with a ratio of 3:1:1.

\noindent \textbf{Gaze Collection} 
Gaze tracks were seamlessly obtained during the bounding box annotation process, utilizing software referenced in~\cite{wang2022follow}.
To enhance focus on cytology images, we refine the software by adding drag and zoom functionalities. The data collection is conducted across five distinct sessions, to ensure the accuracy of annotations and mitigate the impact of experimental fatigue.
~\\
\noindent \textbf{Implementation Details} 
To ensure fairness of the comparison, all experiments are conducted on one NVIDIA RTX 3090 GPU, with an initial learning rate of 3e-4. Our framework is optimized using Adam optimizer with cosine learning rate decay.
CNN-based models are initialized with the default setting provided in MMDetection~\cite{chen2019mmdetection} and trained for 100 epochs.
DETR-like models are initialized by the pre-train weights on CoCo dataset~\cite{lin2014microsoft} and trained for 50 epochs.
Gaze-guided warmup is conducted in the early 10 epochs for CNN-based models and 5 epochs for DETR-like models.
A quantitative evaluation is conducted using average precision (AP) and average recall (AR) as the metrics. 
We calculate the average AP over multiple IoU thresholds from 0.2 to 0.5 with a step size of 0.05 (AP$_{0.2:0.5}$), and individually evaluated AP at the IoU thresholds of 0.2 and 0.5 (denoted as AP$_{0.2}$ and AP$_{0.5}$), respectively.

\subsection{Comparison with the State-of-the-Art}
\begin{table}[ht]
\centering
\caption{Comparison to state-of-the-art methods. Best results are denoted in bold.}
\begin{tabular}{lc|cccc}
\toprule
Method                           & \multicolumn{1}{l|}{Backbone} & $AP_{0.2:0.5}$ & $AP_{0.2}$     & $AP_{0.5}$     & $AR$           \\ \midrule
EllipseNet (MICCAI 2021)\cite{chen2021ellipsenet}          & ResNet-50                      & 0.443          & 0.579          & 0.264          & 0.792          \\
RetinaNet (ICCV 2017)\cite{lin2017focal}             & ResNet-50                      & 0.466          & 0.533          & 0.326          & 0.850          \\
Sparse R-CNN (CVPR 2021) \cite{sun2021sparse}         & ResNet-50                      & 0.386          & 0.553          & 0.149          & 0.685          \\
YOLOv8 (2023)\cite{Jocher_Ultralytics_YOLO_2023}                           & ResNet-50                      & 0.482          & 0.587          & 0.333          & 0.866          \\
DETR (ECCV 2020)\cite{carion2020end}                  & ResNet-50                      & 0.514          & 0.656          & 0.300          & 0.866          \\
Conditional-DETR (ICCV 2021)\cite{meng2021conditional}      & ResNet-50                      & 0.483          & 0.598          & 0.331          & 0.897          \\
DN-DETR (CVPR 2022 Oral)\cite{li2022dn}          & ResNet-50                      & 0.535          & 0.657          & 0.330          & 0.912          \\
DINO (ICLR 2023)\cite{zhang2022dino}                  & Swin-L                        & 0.646          & 0.711          & 0.561          & \textbf{0.988} \\ \midrule
Gaze-DN-DETR                     & ResNet-50                      & 0.573          & 0.697          & 0.348          & 0.893         \\
Gaze-DINO                        & Swin-L                        & \textbf{0.687} & \textbf{0.755} & \textbf{0.634} & \textbf{0.988} \\ \bottomrule
\end{tabular}
\label{tab:comparison}
\end{table}

We first compare our method with the state-of-the-art detection algorithms to evaluate the improved performance using extra help from gaze.
As shown in Table \ref{tab:comparison}, DETR-like models generally show higher precision compared to CNN-based methods. 
Both DN-DETR and DINO exhibit improved performance with the assistance of gaze. Notably, Gaze-DINO achieves the best performance, which demonstrates the efficacy of Gaze-DETR in reducing false positives. 

Qualitative comparisons are also conducted on methods with the backbone of ResNet-50.
As shown in Fig. \ref{fig_result}, both RetinaNet and DN-DETR tend to produce false positive results in areas where experts gaze but do not annotate bounding boxes. 
In contrast, our proposed method can significantly mitigate this situation with the help of expert gaze.


\subsection{Compatibility of our method}
\label{sec:compat}
To demonstrate the compatibility of our method and further validate its effectiveness, we extend our method to other detection models. 

Gaze-guided warm-up can be integrated into CNN-based models seamlessly. A comparative analysis of RetinaNet and Sparse R-CNN, with and without Gaze-guided warm-up, is presented in the first two rows of Table~\ref{tab:com}. The results demonstrate that the incorporation of our method, specifically the inclusion of only gaze-guided warm-up, yields notable performance improvements in CNN-based models.

\begin{table}[ht]
    \centering
    \caption{Results of extending our method to other detection models. Methods with $^\ast$ include only gaze-guided warm-up in the `$+$ Gaze' bar, while others include both gaze-guided warm-up and gaze-guided rectification.}
    \begin{tabular}{lc|cccc|cccc}
\toprule
\multirow{2}{*}{Method}                                                        & \multirow{2}{*}{Backbone} & \multicolumn{4}{c|}{Original}                                     & \multicolumn{4}{c}{$+$ Gaze}                                        \\
                                                                               &                           & $AP_{0.2:0.5}$ & $AP_{0.2}$     & $AP_{0.5}$     & $AR$           & $AP_{0.2:0.5}$ & $AP_{0.2}$     & $AP_{0.5}$     & $AR$           \\ \midrule
RetinaNet$^\ast$                                                                    & ResNet-50                 & 0.466          & 0.533          & 0.326          & 0.850          & \textbf{0.478} & \textbf{0.542} & \textbf{0.332} & \textbf{0.873} \\ \midrule
Sparse R-CNN$^\ast$                                                                   & ResNet-50                 & 0.386          & \textbf{0.553} & 0.149          & 0.685          & \textbf{0.394} & 0.550          & \textbf{0.168} & \textbf{0.692} \\ \midrule
DN-DETR                                                                        & ResNet-50                 & 0.535          & 0.657          & 0.330          & \textbf{0.912}         & \textbf{0.573} & \textbf{0.697} & \textbf{0.348} & 0.893 \\ \midrule
\multirow{2}{*}{\begin{tabular}[c]{@{}l@{}}DN-deformable\\ -DETR\end{tabular}} & ResNet-50                 & 0.488          & 0.595          & 0.354          & 0.969          & \textbf{0.502} & \textbf{0.609} & \textbf{0.367} & \textbf{0.979} \\
                                                                               & ResNet-101                & 0.524          & 0.686          & 0.421          & 0.953          & \textbf{0.568} & \textbf{0.721} & \textbf{0.440} & \textbf{0.976} \\ \midrule
\multirow{2}{*}{DINO}                                                          & Resnet-50                 & 0.481          & 0.576          & \textbf{0.314} & 0.941          & \textbf{0.492} & \textbf{0.589} & 0.310          & \textbf{0.969} \\
                                                                               & Swin-L                    & 0.646          & 0.711          & 0.561          & \textbf{0.988} & \textbf{0.687} & \textbf{0.755} & \textbf{0.585} & \textbf{0.988} \\ \bottomrule
\end{tabular}
    \label{tab:com}
\end{table}
Any DETR-like model can utilize our proposed gaze integration framework. 
We conduct a comparison experiment on different DETR-like models with varying backbones.
The latter part of Table~\ref{tab:com} presents the results, from which
several observations can be drawn: (1) The performance of DETR-like models consistently increases in most cases with the help of expert gaze. (2) Our proposed training strategy is also compatible with other techniques, such as denoise training\cite{li2022dn}, deformable attention\cite{zhu2020deformable}, and query selection\cite{zhang2022dino}, etc.

\subsection{Ablation Study}
We conduct an ablation study on Gaze-DN-DETR with the ResNet-50 backbone and Gaze-DINO with the Swin-L backbone to verify the effectiveness of each component and report the results in Table~\ref{tab:ablation_framework}, from which several observations can be drawn: 
(1) Compared with the baseline, our gaze-guided warm-up achieves considerably better performance.
(2) The gaze-guided rectification strategy alone leads to limited improvements. 
(3) Combining both gaze-guided warm-up and gaze-guided rectification strategies, our method demonstrates the best performance. This is because gaze-guided warm-up introduces an additional category, 'gaze only', in the early stages. This strengthens the classifier's capabilities and reduces the difficulty of gaze-guided rectification.

\begin{table}[t]
    \centering
    \caption{Performance of ablation study for our Gaze-DN-DETR and Gaze-DINO. GGW denotes Gaze-Guided Warm-up and GGR denotes Gaze-Guided Rectification.}
    \begin{tabular}{cc|cccc|cccc}
\toprule
\multicolumn{2}{c|}{Method} & \multicolumn{4}{c|}{DN-DETR (ResNet-50)}                          & \multicolumn{4}{c}{DINO (Swin-L)}                                 \\
+ GGW        & + GGR        & $AP_{0.2:0.5}$ & $AP_{0.2}$     & $AP_{0.5}$     & $AR$           & $AP_{0.2:0.5}$ & $AP_{0.2}$     & $AP_{0.5}$     & $AR$           \\ \midrule
             &              & 0.535          & 0.657          & 0.330          & \textbf{0.912} & 0.646          & 0.711          & 0.561          & \textbf{0.988} \\
\checkmark    &              & 0.565          & 0.685          & 0.330          & 0.880          & 0.672          & 0.749          & 0.584          & 0.981         \\
             & \checkmark    & 0.532          & 0.669          & 0.339          & 0.902          & 0.661          & 0.751          & 0.554          & 0.960          \\
\checkmark    & \checkmark    & \textbf{0.573} & \textbf{0.697} & \textbf{0.348} & 0.893          & \textbf{0.687} & \textbf{0.755} & \textbf{0.634} & \textbf{0.988} \\ \bottomrule
\end{tabular}
    \label{tab:ablation_framework}
\end{table}

\section{Conclusion}
In this study, we propose Gaze-DETR, a novel training strategy that incorporates expert gaze patterns to reduce false positives in the identification of vulvovaginal candidiasis within TCT images.
We seamlessly captured the eye movements of experts using an eye-tracker during data annotation. 
The gaze data offers a window into the diagnostic reasoning of pathologists by highlighting areas scrutinized but not ultimately annotated.
We identified these areas as confounding regions that are prone to false positives for detectors. 
By implementing gaze-guided warm-up and gaze-guided rectification strategies, we effectively harness this supplementary information to enhance the training of detection networks. 
Our experimental results affirm the efficacy and generalizability of Gaze-DETR, paving the way for more accurate medical image analysis through the integration of human expert insight.


\bibliographystyle{splncs04}
\bibliography{ref}
\end{document}